\documentclass[onecolumn, conference, 11pt]{IEEEtran}

 \usepackage[utf8]{inputenc}
\usepackage{blindtext, graphicx}
\usepackage{amsmath, amsfonts, amssymb, bm}
\usepackage{ifpdf}
\usepackage[dvipsnames]{xcolor}
\usepackage[exponent-product=.,detect-weight]{siunitx}
\usepackage[hyphens]{url}
\usepackage[breaklinks]{hyperref}
\hypersetup{hidelinks}

\usepackage[english, plain]{fancyref}

\usepackage[nospace]{cite}
\usepackage{eso-pic}

\newcommand{\bfb}{\mathbf{b}}
\newcommand{\bfA}{\mathbf{A}}
\newcommand{\bfC}{\mathbf{C}}

\newcommand{\bfe}{\mathbf{e}}

\newcommand{\bfH}{\mathbf{H}}

\newcommand{\bfL}{\mathbf{L}}

\newcommand{\bfI}{\mathbf{I}}
\newcommand{\bfK}{\mathbf{K}}

\newcommand{\calM}{\mathcal{M}}
\newcommand{\calN}{\mathcal{N}}

\newcommand{\bfP}{\mathbf{P}}
\newcommand{\bfQ}{\mathbf{Q}}

\newcommand{\bfR}{\mathbf{R}}
\newcommand{\bbR}{\mathbb{R}}

\newcommand{\bfv}{\mathbf{v}}

\newcommand{\bfw}{\mathbf{w}}

\newcommand{\bfx}{\mathbf{x}}

\newcommand{\bfy}{\mathbf{y}}

\newcommand{\by}{\bar{\bfy}}

\newcommand{\boSigma}{\boldsymbol{\Sigma}}

\newcommand{\rchi}{\protect\raisebox{2pt}{$\boldsymbol{\chi}$}}
\newcommand{\bchi}{\hat{\rchi}}
\newcommand{\bochi}{\protect\raisebox{1pt}{$\boldsymbol{\chi}$}}
\newcommand{\hchi}{\hat{\rchi}}
\newcommand{\boomega}{\boldsymbol{\omega}}
\newcommand{\boOmega}{\boldsymbol{\Omega}}
\newcommand{\bomu}{\boldsymbol{\mu}}

\newcommand{\boxi}{\boldsymbol{\xi}}

\newcommand{\bxi}{\bar{\boxi}}

\newcommand{\bfzero}{\mathbf{0}}

\DeclareMathOperator{\col}{col}

\usepackage[english, onelanguage,ruled,vlined, longend, titlenotnumbered]{algorithm2e}
\usepackage{algorithmic}
\usepackage{mathtools}
\usepackage{caption}
\newtheorem{rem}{Remark}
\newtheorem{ex}{Example}
\newtheorem{dfn}{Definition}
\newtheorem{prop}{Proposition}

\usepackage{listings}

\usepackage{tikz,
	pgfplots,
	pgfplotstable,}
\usetikzlibrary{patterns,
	plotmarks}
\pgfplotsset{compat=1.12}
\usepgfplotslibrary{external} 
\begin{document}
\title{A Code for Unscented Kalman Filtering  on   Manifolds (UKF-M)}
\author{\IEEEauthorblockN
    {Martin \textsc{Brossard}\IEEEauthorrefmark{2}, Axel  \textsc{Barrau}\IEEEauthorrefmark{1} and Silv\`ere \textsc{Bonnabel}\IEEEauthorrefmark{2} }
    
    \IEEEauthorblockA{\IEEEauthorrefmark{2}MINES ParisTech, PSL Research University, Centre for Robotics, 60 Boulevard Saint-Michel, 75006, Paris, France}
    
	\IEEEauthorblockA{\IEEEauthorrefmark{1}Safran Tech, Groupe Safran, Rue des Jeunes Bois-Ch\^ateaufort, 78772, Magny Les Hameaux Cedex, France}
}


\maketitle
\begin{abstract}
The present paper introduces a novel methodology for Unscented Kalman Filtering (UKF) on manifolds that extends  previous work by the authors on  UKF on Lie groups. Beyond filtering performance, the main interests of the approach are its versatility, as the method applies to numerous   state estimation problems, and its simplicity of implementation for practitioners not being necessarily familiar with manifolds and Lie groups. We have developed the method on two independent open-source Python and Matlab frameworks we call \emph{UKF-M}, for quickly implementing and testing the approach. The online repositories contain tutorials, documentation, and various relevant robotics examples that the user can readily reproduce and then adapt, for fast prototyping and benchmarking.  The code is available at \texttt{\url{https://github.com/CAOR-MINES-ParisTech/ukfm}}.
\end{abstract}

\section{Introduction}
\label{sec:int}

Over the past fifty years, the Kalman filter has been a pervasive tool in aerospace engineering and beyond, to estimate the state of a system subject to dynamical evolution, see e.g. \cite{barrauInvariant2018}. When the system's dynamics are governed by nonlinear equations, one generally resorts to a  variant called the Extended Kalman Filter (EKF), or to the more recent  Unscented Kalman Filter (UKF) \cite{julierUnscented2004,julierNew1997}. There has been various attempts to adapt the EKF and (respectively)  UKF to the case where the system's state lives in a manifold $\mathcal M$, see respectively \cite{hertzbergIntegrating2013} and \cite{haubergUnscented2013,menegazUnscented2019,loiannoVisual2016,svachaInertial2019}. 

In this paper we introduce \emph{UKF-M}, a novel and general method for UKF on manifolds whose versatility allows direct application  to  numerous manifolds encountered in practice. The theory is supported with independent Python and Matlab open sourced implementations. The framework is well documented, and contains a number of  examples that can be readily run and then adapted, where our methodology spares the analytic computation of Jacobians (contrary to EKF) and is thus well suited to fast prototyping and benchmarking.

Filtering on manifolds is historically motivated by aerospace applications where one seeks to estimate (besides other quantities) the orientation of a body in space. Much work has been devoted to making the EKF work with orientations, namely quaternions or rotation matrices. The idea is  to make the EKF estimate an error instead of the state directly, leading to error state EKFs  \cite{leffensKalman1982, hertzbergIntegrating2013,forbesContinuoustime2014,solaQuaternion2012} and their UKF counterparts\cite{kraftQuaternionbased2003,leeGlobal2016,crassidisUnscented2003}. The set of orientations of a body in space is the Lie group  $SO(3)$ and efforts devoted to estimation on $SO(3)$   have paved the way to EKF on Lie groups, see \cite{bonnabelLeftInvariant2007,barrauIntrinsic2015,bourmaudContinuousDiscrete2015,bourmaudDiscrete2013,barrauInvariant2017,barrauInvariant2018} and unscented Kalman filtering on Lie groups, see 
\cite{bohnUnscented2012,bohnUnscented2016,leeGlobal2016,loiannoVisual2016,brossardUnscented2017,forbesSigma2017,svachaInertial2019}.

Lie groups play a prominent role in robotics \cite{solaMicro2018}. In the context of state estimation and localization, viewing poses as elements of the Lie group $SE(3)$ has proved   relevant  \cite{wangError2006,parkKinematic2008,chirikjianStochastic2009,chirikjianGaussian2014,barfootPose2011,barfootAssociating2014,hartleyContactAided2018}. The use of the novel Lie group $SE_2(3)$ introduced in \cite{barrauInvariant2017} has   led to drastic improvement of  Kalman filters for robot state  estimation   \cite{barrauInvariant2017, koFeatures2018,hartleyContactAided2018, barrauInvariant2018,koImprovement2018,wangGlobally2018,brossardUnscented2018,wuInvariantEKF2017}. Similarly, using   group $SE_k(n)$ introduced for Simultaneous Localization And Mapping (SLAM) in \cite{bonnabelSymmetries2012,barrauEKFSLAM2015} makes EKF  consistent or convergent  \cite{barrauEKFSLAM2015,brossardExploiting2019,heoConsistent2018,heoConsistent2018a,zhangConvergence2017,mahonyGeometric2017}. Finally,  there has been attempts to devise UKFs respecting natural symmetries of the systems' dynamics, namely the invariant UKF, see \cite{condominesNonlinear2013,condominesPiInvariant2014}. 

Besides providing a comprehensive code, our main contribution in terms of methodology is to introduce a novel and general framework for UKF on manifolds that  is simpler than existing methods, and whose versatility allows direct application  to  all manifolds encountered in practice. Indeed,  \cite{loiannoVisual2016,svachaInertial2019} proposes UKF implementations based on the Levi-Civita connection but mastering differential geometry is difficult. \cite{leeGlobal2016,bohnUnscented2012,bohnUnscented2016,loiannoVisual2016} are reserved for $SO(3)$ and $SE(3)$, while 
\cite{forbesSigma2017} is reserved for Lie groups  and requires more knowledge of Lie theory  than the present paper. 
 
In Section \ref{sec:mat}, we introduce a user-friendly approach to UKF on parallelizable manifolds. Section \ref{Lie:sec} applies the approach in the particular case where the manifold is a Lie group and recovers  \cite{brossardUnscented2017}, but without requiring much knowledge of Lie groups. Section \ref{sec:ukfm} describes the open sourced framework. We then show in Section \ref{manifolds:sec} the method may actually be extended to numerous manifolds encountered in robotics. The conclusion section discusses theoretical issues and provides clarifications related to  Kalman filtering on manifolds. 
\color{black}

\section{Unscented Kalman Filtering on Parallelizable Manifolds}
\label{sec:mat}

In this section we describe our simple methodology for UKF on parallelizable manifolds.  Owing to space limitation, we assume the reader to have approximate prior knowledge and intuition about manifolds and  tangent spaces.  

 \subsection{Parallelizable Manifolds}
 
In order to ``write'' the equations of the extended or the unscented Kalman filter on a manifold, it may be advantageous to have   global coordinates for tangent spaces. 

\begin{figure}
    \centering
\includegraphics[width=9cm]{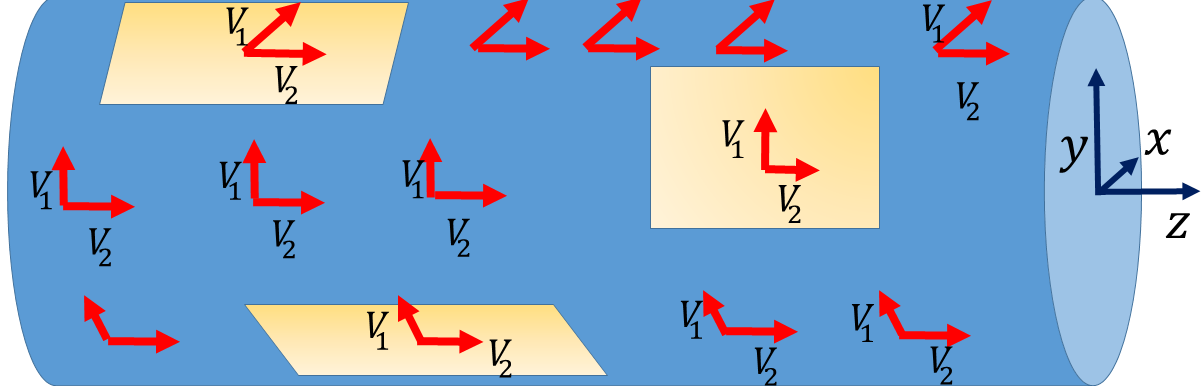}\caption{The cylinder is a parallelizable manifold. We can define vector fields $V_1,V_2$ that form a basis of the tangent space at any point. }\label{cylinder:fig}
\end{figure}

 \begin{dfn}
 A smooth manifold $\mathcal M$ of dimension $d$ is said parallelizable if 
 there exists a set of smooth vector fields $\{V_1,V_2,\cdots,V_d\}$ on the manifold such that for any point $\rchi\in\mathcal M$ the tangent vectors $\{V_1(\rchi),V_2(\rchi),\cdots,V_d(\rchi)\}$ form a basis of the tangent space at $\rchi$. \end{dfn}
 
 \begin{ex}
The cylinder $\{ (x,y,z)\in\mathbb R^3\mid x^2+y^2=1 \}$ is a basic example with $d=2$. $V_1(x,y,z)=(y,-x,0)$ and $V_2=(0,0,1)$  are two tangent vectors that form a local basis at $(x,y,z)$,   see Figure \ref{cylinder:fig}. The cylinder is a simple case but the notion of parallelizable manifolds is much broader. In particular, all 
Lie groups are  parallelizable manifolds.\end{ex}

\begin{ex}\label{ex2}For the rotation matrices $\bfC \in SO(3)$ let us first define the ``wedge'' symbol via
\begin{align}
 \boomega^\wedge=\begin{pmatrix}0&-\omega_3&\omega_2\\\omega_3&0&-\omega_1\\-\omega_2&\omega_1&0\end{pmatrix},\label{wedge}
\end{align} where $\boomega = (\omega_1, \omega_2, \omega_3 )^T$, and choose as vector fields:
\begin{align}
V_1(\bfC)=\bfC \bfe_1^\wedge, ~V_2(\bfC)=\bfC \bfe_2^\wedge, ~V_3(\bfC)=\bfC \bfe_3^\wedge, \label{manif:ea}
\end{align}
where $\bfe_1 = (1,0,0)^T$, $\bfe_2 = (0,1,0)^T$, and $\bfe_3 = (0,0,1)^T$.
\end{ex}
It should be noted, though, that not all manifolds fall in this category. However, we will  see in Section \ref{manifolds:sec} how this issue can be addressed over-parameterizing the state.

\subsection{Uncertainty Representation on Parallelizable Manifolds}
Our  goal is to estimate the state $\rchi \in \mathcal M$ given all the sensor measurements. As sensors are flawed, it is impossible to exactly reconstruct   $\rchi$. Instead, a filter maintains a ``belief" about the state, that is,  its statistical distribution given past sensors' readings. The Kalman filter in $\mathbb R^d$ typically maintains a Gaussian belief such that $ \rchi  \sim \calN\left(\bchi, \bfP\right),$ which may be re-written  in the form:
\begin{equation}
 \rchi =  \bchi +\boxi, \text{~} \boxi \sim \calN\left(\bfzero, \bfP\right). \label{eq:gauss}
\end{equation}
We see that the belief is encoded using only a mean estimate $\bchi$, and a covariance matrix $\bfP$ that encodes the extent of dispersion of the belief around the estimate. 

Consider a parallelizable manifold $\mathcal M$, and let $\{V_1,V_2,\cdots,V_d\}$ denote the associated vector fields. To devise a similar belief on $\mathcal M$, one needs of course local coordinates to write the mean $\bchi\in \mathcal M$. This poses no problem, though. The harder part is to find a way to encode dispersion around the estimate $\bchi$.   It is now commonly admitted that the tangent space at $\bchi$ should encode such dispersion, and that   covariance  $\bfP$ should hence reflect dispersion in the tangent space.  As additive noise \eqref{eq:gauss} makes no sense for $\rchi \in \mathcal M$, we  define a  probability distribution $\rchi \sim \calN_\varphi(\bchi,\bfP)$, for the random variable $\rchi \in \mathcal M$ as 
\begin{equation}
 \rchi = \varphi\left( \bchi,\boxi\right), \text{~} \boxi \sim \calN\left(\bfzero, \bfP\right), \label{eq:left}
\end{equation}
where   $\varphi:\mathcal M\times \bbR^{d}\to\mathcal M$ is a smooth function
 chosen by the user and satisfying $ \varphi\left( \bchi ,
\bfzero\right)=\bchi$.  In \eqref{eq:left}, $ \boxi\in\bbR^d$ is a random Gaussian vector that
encodes directions of the tangent space at $\hchi$, $\calN\left(.,.\right)$ is the
classical Gaussian distribution in Euclidean space, and $\bfP\in \bbR^{d \times
d}$ the associated covariance matrix; and we also impose the Jacobian of $\varphi$ at $( \bchi ,
\bfzero )$ w.r.t. $\boxi$ to be Identity, see \cite{absilOptimization2009}. Using the parallelizable manifold
property, we implicity use  coordinates in the tangent space, as
$\boxi=(\xi^{(1)},\xi^{(2)},\cdots,\xi^{(d)})^T\in\bbR^d$ encodes  the
tangent vector $\xi^{(1)}V_1(\bchi)+ \cdots+\xi^{(d)}  V_d(\bchi)$.  Hence
$\varphi$ is called a ``retraction'', see  \cite{absilOptimization2009}.
In \eqref{eq:left}, the noise-free quantity $\bchi$ is viewed as the mean, and
the dispersion arises  through   $\varphi$. 
We stress that the distribution defined at \eqref{eq:left} is not Gaussian. It is ``only" Gaussian in coordinates related to map $\varphi$. 

\begin{ex}Consider Example \ref{ex2}. Recall tangent vectors at $\bfC$ indicate small motions around $\bfC\in SO(3)$. Tangent vector $\bfC \boomega^\wedge$ indeed writes $\omega_1V_1(\bfC)+\omega_2V_2(\bfC)+\omega_3V_3(\bfC)$, see \eqref{manif:ea}. We can then choose for $\varphi$ the following $\varphi(\bfC,\boomega)=\bfC\exp\left(\boomega^\wedge\right)$, with $\exp$ the exponential map on $SO(3)$. \label{ex:omega}\end{ex}

Finding an appropriate map $\varphi$ is not always straightforward. However there exists in theory some ``canonical'' $\varphi$.
\begin{prop}One may define define $\varphi(\bchi,\boxi)$ as the point of $\mathcal M$ obtained by starting from $\bchi$ and integrating the vector field $\sum_{i=1}^d\boxi^{(i)}V_i$ during one unit of time.  In that case we call $\varphi$ an ``exponential map".\label{rem:exp}\end{prop} However, we sometimes have no closed form for the exponential map and one resorts to simpler retractions $\varphi$. 

\subsection{Bayesian Estimation  Using the Unscented Transform}
\label{sec:bay}

Consider a random variable $\rchi \in \calM$ with  prior probability distribution $p\left(\rchi\right)$. Suppose  we obtain some additional information about $\rchi$ through a measurement $\bfy$. The goal  is to compute   the posterior distribution $p(\rchi|\bfy)$. 
Let
\begin{align}
   &\bfy= h\left(\rchi\right) + \bfv \label{eq:ymesure},
\end{align}
be a  measurement, where $h(.):\calM\rightarrow \bbR^p$ represents the observation function and $\bfv\sim \calN(\bfzero,\bfR)$ is a white Gaussian noise in $\bbR^p$ with known characteristics.
The problem of Bayesian estimation we consider is as follows:
\begin{enumerate}
 \item assume the prior distribution to follow (\ref{eq:left})  with known parameters $\bchi$ and $\bfP$;
 \item assume one measurement $\bfy$ of  (\ref{eq:ymesure}) is available;
 \item approximate the posterior distribution as \begin{equation}
 p(\rchi|\bfy)\approx\varphi(\bchi^+,\boxi^+),\label{eq:pos}\end{equation} where $\boxi^+\sim \mathcal N(\bfzero,\bfP^+)$, and  find parameters $\bchi^+$ and $\bfP^+$.
\end{enumerate}
Letting $\rchi= \varphi\left( \bchi , \boxi\right)$  in \eqref{eq:ymesure}, we see $\bfy$  provides an information about $\boxi \sim \calN\left(\bfzero, \bfP\right)$ and we may use the unscented transform  of \cite{julierNew1997,julierUnscented2004} to  approximate the posterior $p(\boxi|\bfy)$  for $\boxi$ as follows, see Algorithm \ref{alg:1}: we compute a finite number of samples $\boxi_j  $, $j=1,\ldots, 2d$, and pass each of these so-called sigma points through the measurement function
\begin{equation}
 \bfy_j= h\left(\varphi(\bchi ,  \left(\boxi_j\right) \right)   , ~j=1,\ldots,2d \label{eq:yj}.
\end{equation} 
By noting $\bfy_0= h(\varphi(\bchi ,\bfzero))$ we then compute successively the measurement mean $\by = w_m\bfy_0 + \sum_{j=1}^{2d}w_j \bfy_j$, the measurement covariance $\bfP_{\bfy \bfy} = \sum_{j=0}^{2d}w_j (\bfy_j - \by) (\bfy_j - \by)^{T}+\bfR$ and the cross-covariance $\bfP_{\boxi \bfy} = \sum_{j=1}^{2d}w_j \boxi_j \left(\bfy_j - \by\right)^{T}$, where $w_m$ and $w_j$ are weights defined in \cite{brossardUnscented2017,julierNew1997} (see definition of scale parameter $\lambda$  therein also). We then derive the conditional distribution of $\boxi \in \mathbb R^d$ as
\begin{align}
 p(\boxi|\bfy) \sim \calN\left(\bxi,\bfP^+\right), \text{~where}\label{eq:sim} 
 \end{align}
 \begin{align}
  \bfK = \bfP_{\boxi \bfy}\bfP_{\bfy \bfy}, ~\bxi  = \bfK\left(\bfy- \by\right),~ \bfP^{+} = \bfP- \bfK\bfP_{\bfy \bfy}\bfK^T. \label{eq:exbchi} 
\end{align}
This may be viewed as a Kalman update on the error $\boxi$, in the vein of error state Kalman filtering, see e.g. \cite{solaQuaternion2012}. 
The problem is then to convert this into a distribution on the manifold in the form \eqref{eq:left}. We first represent  $p(\boxi|\bfy)$ as $\bar\boxi+\boxi^+$ with $\boxi^+\sim \calN\left(\bfzero,\bfP^+\right)$ and $\bar\boxi$ considered as a   noise free mean.  We suggest to define the posterior $p(\rchi|\bfy)$ as 
\begin{equation}
\rchi\approx \varphi (\bchi^{+},\boxi^+),\quad  \boxi^+\sim \calN\left(\bfzero,\bfP^+\right),
\label{eq:yup2}
\end{equation} where we have let
\begin{equation}
  \bchi^{+} =\varphi  \left( \bchi  , \bxi\right).\label{eq:yup}
\end{equation}

\begin{algorithm}[t]
	\small
	\KwIn{$\bchi, \bfP, \bfy, \bfR$\;}
	\tcp{set sigma points}  
	\nl $\boxi_j =   \col(\sqrt{(\lambda +d)\bfP})_{j}, ~j=1,\ldots,d$,\\
	$\boxi_j = - \col(\sqrt{(\lambda +d)\bfP})_{j-d}, ~j=d+1,\ldots,2d$\;
	\tcp{compute measurement sigma points} 
	\nl $\bfy_0= h(\varphi(\bchi ,\bfzero))$\; 
	\nl $\bfy_j= h(\varphi(\bchi ,\boxi_j)) , ~j=1,\ldots,2d$\;
	\tcp{infer covariance matrices} 
	\nl $\by = w_m\bfy_0 + \sum_{j=1}^{2d}w_j \bfy_j$\;
	\nl $\bfP_{\bfy \bfy} = \sum_{j=0}^{2d}w_j (\bfy_j - \by) (\bfy_j - \by)^{T}+\bfR$\;
	\nl $\bfP_{\boxi \bfy} = \sum_{j=1}^{2d}w_j \boxi_j \left(\bfy_j - \by\right)^{T}$\;
	
	\tcp{update state and covariance} 
	\nl $\bfK = \bfP_{\boxi \bfy}\bfP_{\bfy \bfy}^{-1}$ \tcp*[r]{gain matrix}
	\nl $\bchi^{+} = \varphi (\bchi,\bfK(\bfy- \by))$\;
	\nl $\bfP^{+} = \bfP- \bfK\bfP_{\bfy \bfy}\bfK^T$\;
	\KwOut{$\bchi^+, \bfP^+$\;}
	\caption{Bayesian updating on parallelizable manifolds with prior \eqref{eq:left} and observation \eqref{eq:ymesure}\label{alg:1}}
\end{algorithm}

Note  the approximation done in  \eqref{eq:yup2}-\eqref{eq:yup} actually consists in writing $\varphi\left(\bchi, \bxi+\boxi^+\right)\approx \varphi\left(\varphi\left (\bchi, \bxi\right),\boxi^+\right)$.

 When $\mathcal M=\bbR^d$  the latter equality holds up to the first order in the dispersions $\bxi$, $\boxi^+$, both assumed small.
 In the case where $\mathcal M$ is not a vector space, it may be geometrically interpreted as saying that moving from $\bchi$ along the direction $\bxi+\boxi^+$ approximately consists in moving from $\bchi$ along $\bxi$ and then from the obtained point on $\mathcal M$ along $\boxi^+$.

\subsection{Unscented Kalman Filtering on Parallelizable Manifolds}
\label{sec:uns}
 Consider the dynamics
\begin{eqnarray}
  \rchi_{n} = f\left(\rchi_{n-1},\boomega_n ,  \bfw_n\right), \label{eq:sys} 
\end{eqnarray}
where the state $\rchi_n$ lives in a parallelizable manifold $\mathcal M$, $\boomega_n$ is a known input variable and $\bfw_n \sim \calN\left(\bfzero, \bfQ_n \right)$ is a white Gaussian noise in $\bbR^q$. We consider  observations of the form
\begin{align}
   \bfy_{n}= h\left(\rchi_n  \right) + \bfv_n \label{eq:ymesure2},
\end{align}
where $\bfv_n\sim \calN(\bfzero,\bfR_n)$ is a white Gaussian noise with known
covariance that we assume additive for clarity of the algorithm derivation only. 
For system \fref{eq:sys}-\eqref{eq:ymesure2}, we  model the state posterior conditioned on past measurements using the uncertainty representation \eqref{eq:left}.  To propagate the state, we start from the prior distribution 
$
  p\left(\rchi_{n-1}\right) 
\sim \varphi(\bchi_{n-1},\boxi_{n-1})$ with 
$\boxi_{n-1} \sim \mathcal N (\bfzero,\bfP_{n-1})$
and $\bchi_{n-1}$,  $\bfP_{n-1}$ known, and we seek to compute the state propagated distribution in the form
\begin{equation}
 p\left(\rchi_{n}|\rchi_{n-1}\right) \sim  \varphi(\bchi_{n},\boxi_{n})\quad\text{with}\quad \boxi_{n} \sim \mathcal N (\bfzero,\bfP_n).
\end{equation}

We define sigma points using \eqref{eq:left} and the statistics of noise $\bfw_n$,  and   pass them through  \eqref{eq:sys}. Then, to find $\bchi_n$ one  is faced with the optimization problem of computing a weighted mean on  $\mathcal M$. This route has already been advocated in \cite{kraftQuaternionbased2003,leeGlobal2016,crassidisUnscented2003,forbesSigma2017}. However, to keep the implementation simple and analog to the EKF, we suggest to merely propagate the mean using the unnoisy state model, leading to
\begin{equation}
 \bchi_{n} = f(\bchi_{n-1}, \boomega_{n}, \bfzero).\label{eq:chiprop}
\end{equation}

To compute  the covariance $\bfP_n$ from   $\bfP_{n-1}$  of $\boxi_{n-1}$ we use the fact $\bfw_{n}$ and $\boxi_{n-1}$ are uncorrelated and proceed in two steps. $\mathbf{1)}$ we generate sigma points in $\bbR^d$ corresponding to $\bfP_{n-1}$ and pass them through the unnoisy model \eqref{eq:chiprop} for  nonlinear propagation of $\bfP_{n-1}$ through $f$. We obtain points $\rchi_{n}^j$ on the manifold $\mathcal M$, and  the distribution of propagated state is described as   $\varphi\left(\bchi_{n} ,\boxi_n\right) $, with $\bchi_n$ known from \eqref{eq:chiprop}. We   need to be able to locally invert $\boxi\mapsto\varphi(\bchi,\boxi)$, i.e., to find a map   denoted  by $\varphi^{-1}_{\bchi}(\cdot):\mathcal M\to\bbR^d$  such that 
\begin{equation}\varphi^{-1}_{\bchi}\left(\varphi(\bchi,\boxi)\right)=\boxi+O(||\boxi||^2),\end{equation}that is, a map that allows one to assess the discrepancy between $\bchi$ and $\varphi(\bchi,\boxi)$ is $\boxi$ indeed. Then we      use  $ \varphi^{-1}_{\bchi_{n}}$ to map  sigma points $\rchi_{n}^j$ back into $\bbR^d$ and compute their empirical  covariance $\boSigma_n$. $\mathbf{2)}$ we then generate sigma points for process noise $\bfw_n$ similarly and obtain another covariance matrix encoding dispersion in $\bbR^d$ owed to noise, that adds up to  $\boSigma_n$ and thus clearly distinguish the contribution of the state error dispersion $\boxi_n$ from  noise $\bfw_n$. 
When a new measurement arrives, belief is updated via  Algorithm \ref{alg:1}. 
Algorithm \ref{alg:2} summarizes both  steps, where the weights defined through $\mathtt{set\_weights}(d, \alpha)$ depend  on a scale parameter $\alpha$ (generally set between $10^{-3}$ and $1$), and sigma point dimension, see \cite{brossardUnscented2017,julierNew1997} and documentation in source code.

\begin{algorithm}[t]
	\small
 \KwIn{$\bchi_{n-1}, \bfP_{n-1}, \boomega_{n}, \bfQ_{n}, \bfy_{n}, \bfR_n, \alpha$\;}
 \SetKwBlock{Propagation}{Propagation}{end}
 \Propagation{
 	\tcp{propagate mean state}
  \nl $\bchi_{n} = f(\bchi_{n-1}, \boomega_{n}, \bfzero)$\;
  \tcp{propagate state error covariance}
  \nl $\lambda_, \{w_j\}_{j=0,\ldots,2d} = \mathtt{set\_weights}(d, \alpha)$\;
  \nl $\boxi_j =   \col(\sqrt{(\lambda +d)\bfP_{n-1}})_{j}, ~j=1,\ldots,d$,\\
  $\boxi_j = - \col(\sqrt{(\lambda +d)\bfP_{n-1}})_{j-d}, ~j=d+1,\ldots,2d$\;
  \tcp{use retraction onto manifold}
  \nl $\bochi^j_{n}= f(\varphi(\bchi_{n-1} ,\boxi_{j}),\boomega_{n}, \bfzero), j=1,\ldots,2d$\;
  \tcp{inverse retract to go back in  $\bbR^{d}$}
  \nl $\boSigma_n= \sum_{j=1}^{2d}w_j \varphi^{-1}_{\bchi_{n}}(\bochi_{n}^j)\left(\varphi^{-1}_{\bchi_{n}}(\bochi_{n}^j)\right)^T$\;
  \tcp{proceed similarly for noise}
\nl $\lambda, \{w_j\}_{j=0,\ldots,2q} = \mathtt{set\_weights}(q, \alpha)$\;
\nl $\bfw^j =   \col(\sqrt{(\lambda +q)\bfQ_n})_{j}, ~j=1,\ldots,q$,\\
$\bfw^j = - \col(\sqrt{(\lambda +q)\bfQ_n})_{j-d}, ~j=q+1,\ldots,2q$\;
\nl $\tilde\bochi^j_{n}= f(\bchi_{n-1},\boomega_{n}, \bfw^j), j=1,\ldots,2q$\;
\nl $\bfP_n = \boSigma_n + \sum_{j=1}^{2q}w_j \varphi^{-1}_{\bchi_{n}}(\tilde\bochi_{n}^j)(\varphi^{-1}_{\bchi_{n}}(\tilde\bochi_{n}^j))^T$\;

  }
 \SetKwBlock{Update}{Update (when measurement $\bfy_n$ arrives)}{end}
 \Update{Compute $\bchi_n^+, \bfP_n^+$ from Algorithm \ref{alg:1} with $\bchi_n, \bfP_n$\;
 }
  \KwOut{$\bchi_n^+, \bfP_n^+$\;}
   \caption{UKF on parallelizable manifolds} \label{alg:2}
\end{algorithm}

Using \eqref{eq:chiprop} to  propagate  the  mean  while using  sigma  points  to compute covariance  is also done in  \cite{barfootAssociating2014}, in the particular case of pose compounding on $SE(3)$, with $\varphi$ the $SE(3)$ exponential map.

 \section{Application to UKF on Lie Groups}\label{Lie:sec}
 
To apply the preceding methodology to any $d$-dimensional group $G=\mathcal M$,  one   first defines a basis of the Lie algebra. Then, to any vector $\boxi\in\bbR^d$, one may associate an element denoted by $\boxi^\wedge$ of the Lie algebra $\mathfrak g$. Let  the vee operator $\vee$ denote its inverse, as in e.g., \cite{barfootAssociating2014}. The Lie exponential map ``$\exp$'' maps elements of the Lie algebra to the group. In \eqref{eq:left} we may choose $\varphi(\bchi,\boxi):=\bchi\exp(\boxi^\wedge)$, 
 which corresponds to left concentrated Gaussians on Lie groups \cite{bourmaudDiscrete2013}. Note that, 
in the Lie group case, choosing left invariant vector fields for the $V_i$'s and following Proposition \ref{rem:exp} we exactly recover the latter expression. 

We may invert $\varphi$ using the logarithm map $\exp^{-1}:=\log$ of $G$, and we get 
 \begin{equation}
  \varphi(\bchi,\boxi):=\bchi\exp(\boxi^\wedge), ~ \varphi^{-1}_{\bchi}(\rchi):=\log\left(\bchi^{-1}\rchi\right).\label{ll:eq}
\end{equation} If we  alternatively  privilegiate right multiplications  we have
 \begin{equation}
 \varphi(\bchi,\boxi):=\exp(\boxi^\wedge)\bchi,~ \varphi^{-1}_{\bchi}(\rchi):=\log\left(\rchi\bchi^{-1}\right).
\label{rr:eq}
\end{equation}


 \subsection{Applications in Mobile Robotics: the Group $SE_k(d)$}\label{code:sec}
 
 It is well known that  orientations of body in spaces are described by elements of $SO(3)$. 
It is also well known that the use of $SE(3)$ is advantageous to describe the position and the orientation of a robot (pose), especially for estimation, see    \cite{wangError2006,parkKinematic2008,chirikjianStochastic2009,chirikjianGaussian2014,barfootPose2011,barfootAssociating2014,hartleyContactAided2018}.   In  \cite{barrau2015non, barrauInvariant2017} the group of double direct isometries $SE_2(3)$ was introduced to address estimation problems for robot navigation when the motion equations are based on an Inertial Measurement Unit (IMU). In  \cite{bonnabelSymmetries2012,barrauEKFSLAM2015} the group of multiple spatial isometries $SE_k(d)$ was introduced in the context of SLAM. The group $SE_k(d)$,   allows recovering $SE(3)$ with $k=1,d=3$, $SE(2)$ with $k=1, d=2$  and $SO(3)$ with $k=0,d=3$. It seems to cover virtually all robotics applications where the Lie group methodology has been so far useful (along with trivial extensions to be mentioned in Section \ref{mixed:sec}). Since it was introduced for navigation and SLAM, this group has   been successfully used in various contexts, see \cite{barrauInvariant2017, koFeatures2018,hartleyContactAided2018, barrauInvariant2018,koImprovement2018,wangGlobally2018,brossardUnscented2018,wuInvariantEKF2017,barrauEKFSLAM2015,brossardExploiting2019,heoConsistent2018,heoConsistent2018a,zhangConvergence2017,mahonyGeometric2017,wang2018geometric}.  For more information  see the  code documentation.

 \lstset{
 	language=Python, 
 	basicstyle=\ttfamily\small,
 	numberstyle=\footnotesize,
 	tabsize=2,
 	title=\lstname,
 	escapeinside={\%*}{*)},
 	breaklines=true,
 	frame=lines,
 	breakatwhitespace=true,
 	extendedchars=false,
 	inputencoding=utf8,
 	keywordstyle=\color{blue},
 	stringstyle=\color{red},
 	commentstyle=\color{ForestGreen}
 }

 \subsection{The Mixed Case}\label{mixed:sec} We call mixed the case where $\mathcal M=G\times \bbR^N$. This typically arises when one wants to estimate some additional parameters besides the state assumed to live in the group $G$, such as sensor biases.   By decomposing the state as $\bchi=(\bchi_1,\bchi_2)\in G\times \bbR^N$ and letting $\boxi=(\boxi_1,\boxi_2)$, we typically define $\varphi$ through right multiplication as 
\begin{align}
\varphi\left( \bchi,\boxi\right)=(\exp \left(\boxi_1\right)\bchi_1 ,\bchi_2+\boxi_2) \label{eq:mixed}
\end{align}or if left multiplications are privilegiated  
 $
\varphi\left( \bchi,\boxi\right)=(\bchi_1\exp \left(\boxi_1\right) ,\bchi_2+\boxi_2).$ This way, as many additional quantities as desired may be estimated along the same lines. 
\begin{rem}When $G=SE(3)$ for example, it is tempting   to let $G'=SO(3)$ and to treat $SE(3)$ as $SO(3)\times \bbR^3$ along the lines of mixed systems.  However, in  robotics contexts, it has been largely argued   the Lie group structure of $SE(3)$ to treat poses is more relevant than $SO(3)\times \bbR^3$, as accounting for the coupling between orientation and position leads to important properties, see \cite{wangError2006,parkKinematic2008,chirikjianStochastic2009,chirikjianGaussian2014,barfootPose2011,barfootAssociating2014,hartleyContactAided2018}. In the same way,   $SE_k(3)$ resembles $SO(3)\times \bbR^{3k}$ but has a special noncommutative group structure having recently led to many successes in robotics, see \cite{barrauEKFSLAM2015,barrauInvariant2017, koFeatures2018,hartleyContactAided2018,koImprovement2018,barrauInvariant2018,wangGlobally2018,brossardUnscented2018,wuInvariantEKF2017,brossardExploiting2019,heoConsistent2018,heoConsistent2018a,zhangConvergence2017,mahonyGeometric2017,wang2018geometric}.  \end{rem}
 
\begin{ex}\label{ex3}
	The state $\rchi$ for fusing IMU with GNSS may be divided into the vehicle state $\rchi_1 \in SE_2(3)$ (orientation, velocity and position of the vehicle) and  IMU biases $\rchi_2 =\bfb \in \bbR^6$, see e.g. our example on the KITTI dataset \cite{geigerVision2013}. Further augmenting $\rchi_2$ with new parameters, e.g. time synchronization and force variables  \cite{nisarVIMO2019}, is straightforward.
\end{ex}

\section{UKF-M Implementation} \label{sec:ukfm}

We have released both open source Python package and Matlab toolbox \emph{UKF-M} implementations of our method at \texttt{\url{https://github.com/CAOR-MINES-ParisTech/ukfm}}. Both implementations are wholly independent,  and their design guidelines pursue simplicity, intuitiveness and easy adaptation  rather than optimization. We adapt  the code to the user preferences as follow: the Python code follows class-object paradigm and is heavily documented through the Sphinx documentation generator, whereas the Matlab toolbox contains equivalent functions without class as we believe choosing well function names is best suited for the Matlab use as compared to  class definition. The following code snippets are based on the Python package that we   recommend using.
\subsection{Recipe for Designing a UKF on Manifolds}\label{sec:recipe}

To devise an UKF for any fusion problem on a parrallelizable manifold (or Lie group) $\calM$   the ingredients required in terms of implementation are as follows, see Snippet 1.

\lstset{language=Python}
\begin{figure}
\begin{lstlisting}[title=Snippet 1: how to devise an UKF on manifolds, morekeywords={UKF, ukfm},]
ukf = ukfm.UKF(
f=model.f,             # propagation model
h=model.h,             # observation model
phi=user.phi,          # retraction
phi_inv=user.phi_inv,  # inverse retraction 
Q=model.Q,             # process  cov.
R=model.R,             # observation  cov.
alpha=user.alpha       # sigma point param.
state0=state0,         # initial state
P0=P0)                 # initial covariance
\end{lstlisting}
\begin{lstlisting}[title={Snippet 2: setting $\varphi$, $\varphi^{-1}$ for $\rchi:=\left(\mathtt{Rot} \in SO(3), \mathtt{v}, \mathtt{p}\right)$}]
def phi(state, xi):
return STATE(
Rot=state.Rot.dot(SO3.exp(xi[0:3])),
v=state.v + xi[3:6]
p=state.p + xi[6:9])

def phi_inv(state, hat_state):
return np.hstack([ # concatenate errors
SO3.log(hat_state.Rot.T.dot(state.Rot)),
state.v - hat_state.v,
state.p - hat_state.p])
\end{lstlisting} 

\end{figure}
\begin{enumerate}
	\item A model that specifies the functions $f$ and $h$ used in the filter;
	\item An uncertainty representation \eqref{eq:left}. This implies an expression for the function $\varphi$ and its inverse $\varphi^{-1}$,  defined by the user;
	\item Filter parameters, that define noise covariance matrices $\bfQ_n$, $\bfR_n$ and weights ($\lambda$, $w_m$, and $w_j$) through $\alpha$. Noise covariance values are commonly guided by the model and tuned by the practitioner, whereas $\alpha$ is generally set between $10^{-3}$ and $1$ \cite{julierNew1997}.
	\item Initial state estimates $\bchi_0$ and $\bfP_0$.
\end{enumerate}

\begin{ex}
	Consider a 3D model whose state contains a rotation matrix $\mathtt{Rot} \in SO(3)$, the velocity $\mathtt{v} \in \bbR^3$ and position $\mathtt{p} \in \bbR^3$ of a moving vehicle. Defining $\varphi$ and $\varphi^{-1}$ allows   computing (respectively) a new state and a state error. One possibility is given in Snippet 2, where $\rchi \in SO(3)\times \bbR^6 $, $\varphi(\hchi,\boxi) = \left(\mathtt{\hat{Rot}}\exp(\boxi^{(0:3)}), \hat{\mathtt{v}}+\boxi^{(3:6)}, \hat{\mathtt{p}}+\boxi^{(6:9)}\right)$ and $\varphi^{-1}_{\hchi}(\rchi) = (\log(\mathtt{\hat{Rot}}^{T}\mathtt{Rot}), \mathtt{v}-\hat{\mathtt{v}}, \mathtt{p}-\hat{\mathtt{p}})$. \label{ex:martin}
\end{ex}

In the particular case where $\mathcal M$ is a Lie group  we follow the rules above but we simplify step 2) as follows: we pick an uncertainty representation, either \eqref{ll:eq} or  \eqref{rr:eq}. This directly implies an expression for the map $\wedge$ and its inverse $\vee$, as well as for the exponential $\exp$ and its (local) inverse $\log$. Applying the present general methodology for the particular case of Lie groups, we  recover the method of \cite{brossardUnscented2017}. 

\begin{ex}
	We may modify the representation used in Example \ref{ex:martin} by viewing the state as an element $\rchi \in SE_2(3)$ instead. This defines two alternative retractions. See e.g.  implementation for corresponding $\varphi^{-1}$'s in Snippet 3. A quick comparison displayed in Figure \ref{fig:nav} indicates  the $SE_2(3)$-UKF with right multiplications \eqref{rr:eq} outperforms the other filters, notably the one based on the naive structure of Example \ref{ex:martin}.
\end{ex}

\subsection{Implemented Examples}
In the code, we implement the frameworks on relevant vanilla robotics examples which are listed as follows:
\begin{itemize}
	\item 2D vanilla robot localization tutorial based on odometry and GNSS measurements;
	\item 3D attitude estimation from an IMU equipped with gyro, accelerometer and magnetometer;
	\item 3D inertial navigation on flat Earth where the vehicle obtains observations of known landmarks;
	\item 2D SLAM where the UKFs follows \cite{huangQuadraticComplexity2013} to limit computational complexity and adding new observed landmarks in the state;
	\item IMU-GNSS fusion on the KITTI dataset \cite{geigerVision2013};
	\item an example where the state lives on the 2-sphere manifold, modeling e.g., a spherical  pendulum \cite{kotaruVariation2019}.
\end{itemize}

\begin{figure}
\begin{lstlisting}[title={Snippet 3: defining $\varphi^{-1}$ via \eqref{ll:eq} or \eqref{rr:eq} for $\rchi \in SE_2(3)$}]
def phi_inv(state, hat_state):
  chi = state2chi(state)
  hat_chi = state2chi(hat_state)
  # if left multiplication (17)
  return SEK3.log(SEK3.inv(hat_hat).dot(chi))
  # if right multiplication (18)
  return SEK3.log(chi.dot(SEK3.inv(hat_hat)))
\end{lstlisting}

\end{figure}

We finally enhance code framework, documentation and examples with filter performance comparisons: for each  example we simulate   Monte-Carlo data and benchmark   UKFs and EKFs  based on different choices of   uncertainty representation \eqref{eq:left}  through accuracy and consistency metrics.  

 \begin{ex}
 	Figure  \ref{fig:nav} displays two EKFs and two UKFs for inertial navigation in the setting of \cite{barrauInvariant2017}, where initial heading and position errors are large, respectively 45 degrees and 1 m. The second UKF, whose uncertainty representation \eqref{eq:left}  is based on $SE_2(3)$ exponential, see Section \ref{code:sec}, clearly outperforms the EKF, the first UKF, and improves the EKF of \cite{barrauInvariant2017} during the first 10 seconds of the trajectory.
 \end{ex}

\section{Extension to General Manifolds}\label{manifolds:sec}

\begin{figure}
	\centering
	\includegraphics[width=9cm]{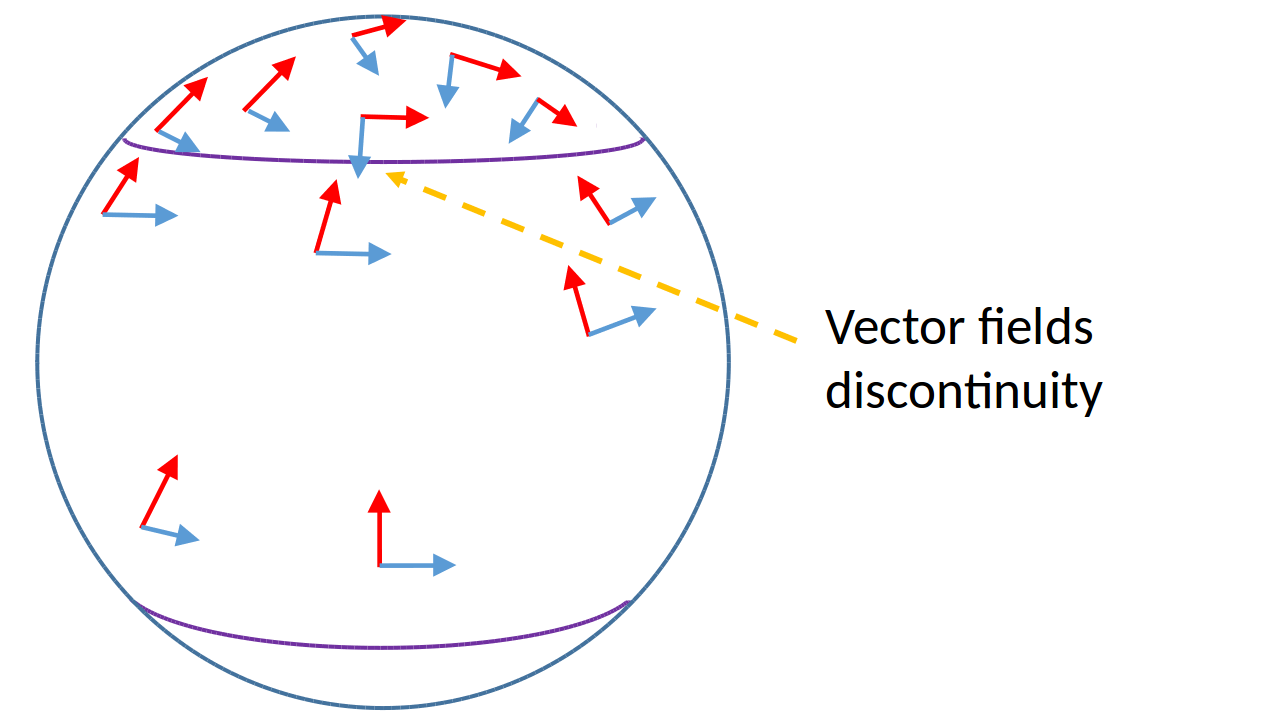}\caption{  We see covering the 2-sphere with 3 parallelizable patches (in between polar circles, and beyond each) inevitably induces discontinuities that may degrade filtering performances. This is a consequence of the  theorem that states it is not possible to ``comb a hairy ball'', see \cite{milnor1978analytic}. }\label{sphere:fig} 
\end{figure}

The main problem when $\mathcal M$ is not parallelizable is that one cannot define a global uncertainty representation through a map $\varphi$ as in \eqref{eq:left}. Indeed $\boxi=(\boxi^{(1)},\cdots,\boxi^{(d)})$ encodes  at any $\rchi\in\mathcal M$ coordinates in the tangent space related to a basis $(V_1(\rchi),\cdots,V_d(\rchi))$ of the tangent space. 
 On general manifolds, though,  it is always possible to cover  the manifold with ``patches" $\mathcal M_1,\cdots,\mathcal M_K$, such that on each patch $i$ we have a set of vector fields $(V_1^{(i)},\cdots,V_d^{(i)})$ allowing one to apply our methodology. For instance on the 2-sphere one could choose a North-East frame in between the polar circles, and then some other smooth set of frames beyond polar circles. However two main issues arise. First, we feel such a procedure induces discontinuities at the polar circles that will inevitably degrade the filter perfomances. Indeed  by moving $\bchi$ slightly at the polar circle, one may obtain a jump in the distribution $\mathcal N_\varphi(\bchi,\bfP)$ with fixed covariance $\bfP$, see Figure \ref{sphere:fig}. Then, we see the obtained filter wholly depends on the way patches are chosen, which is undesirable.

\subsection{The Lifting ``Trick''}\label{trick:sec}

It turns out a number of manifolds of interest called homogeneous spaces   may be ``lifted''  to a Lie group, hence a parallelizable manifold. By simplicity\footnote{Generalizations to th\label{key}e Stiefel manifold $St(p,n)$, that is, a set of $p$ orthonormal vectors of $\bbR^n$,  and hence to the  set of $p$-dimensional subspaces of $\bbR^n$ called the Grassmann manifold are then  straightforward. } we  consider  as a tutorial example the 2-sphere $\mathcal M=\mathbb S^{2}=\{\bfx\in\bbR^3\mid ||\bfx||=1\}$ with state $\bfx_n\in\mathbb S^{2}$. As $\bfx_{n+1}$ and $\bfx_n$ necessarily lie on the sphere, they are related by a rotation, that is,
\begin{equation}\bfx_{n+1}=\boOmega_n \bfx_n\label{sphere:dyn:eq}\end{equation}
with $\boOmega_n\in SO(3)$ that may be written as $\exp(\boomega_n^\wedge)\exp(\bfw_n^\wedge)$ where $\boomega_n$ is a known input, and $\bfw_n\sim\mathcal N(0,\bfQ_n)$ represents a noise, see \eqref{wedge} for the definition of wedge operator, and $\exp$ is the usual matrix exponential of $SO(3)$. We assume $\bfx_n$ is measured through a linear observation, that is,
\begin{align}
\bfy_n=\bfH\bfx_n +\bfv_n\in\bbR^p.
\end{align}

\begin{figure}
	\centering 	
	\includegraphics{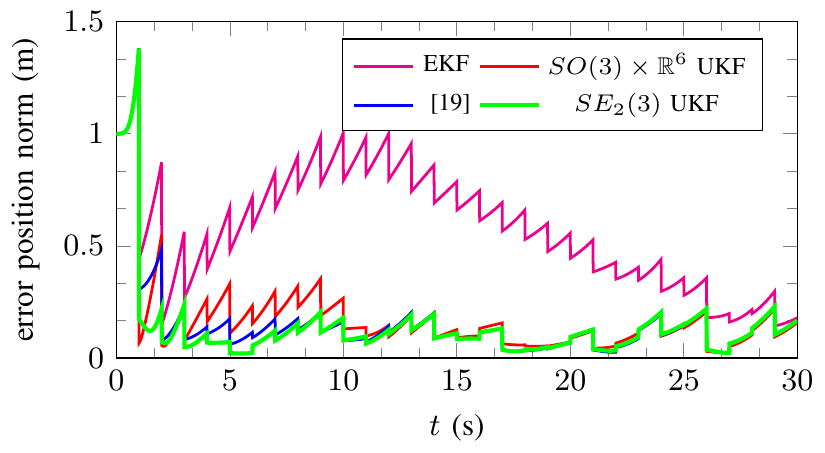}

	\caption{Inertial navigation with heavy initial errors in the setting of \cite{barrauInvariant2017}. $SE_2(3)$-UKF obtains the best results. \label{fig:nav}}
	
\end{figure}

\begin{ex} We provide a (novel) script which simulates a point of a pendulum with stiff wire living on a sphere, where two components are measured through e.g. a monocular camera, i.e. $\bfH = [\bfe_1, \bfe_2]^T $. 
\end{ex}

The dynamics can be lifted into $SO(3)$ by writing $\bfx_n$ via a  rotation matrix $\bfR_n$, that is, we posit $\bfx_n=\bfR_n\bfL$ with $\bfL\in\mathbb R^3$. In terms of $\bfR_n$, dynamics \eqref{sphere:dyn:eq} may be lifted letting $\bfR_{n+1}=\boOmega_n \bfR_n$ as then $\bfR_n\bfL$ satisfies \eqref{sphere:dyn:eq} indeed. Similarly, the output in terms of $\bfR_n$ writes $\bfy_n=\bfH\bfR_n\bfL +\bfv_n=\tilde h(\bfR_n)+\bfv_n$.  Having transposed the problem into estimation on the parallelizable manifold $SO(3)$, we can then apply the two UKFs by setting $\varphi$  to either \eqref{ll:eq} or \eqref{rr:eq}.

     \subsection{Covariance Retrieval}
     
The practitioner may wonder how to retrieve the covariance in the original variables.   Assume we have a Gaussian vector $\bfx\sim\mathcal (\bomu,\boSigma)$, and we want to approximate $g(\bfx)$ as a Gaussian. This might addressed resorting to the unscented transform but a more basic and direct approach is as follows.
Consider $\bfA$ a matrix and $\bfb$ a vector. Then it is known from probability theory that  \begin{align}\bfA \bfx+\bfb \sim\mathcal (\bfA\bomu+ \bfb ,\bfA\boSigma \bfA^T).\label{gauss:t} \end{align}  
Then, we can write $\bfx=\bomu+ \bfe$ with $e\sim \mathcal N(\bfzero, \boSigma)$ and linearizing we find 
     $
     g(\bfx)\approx g(\bomu)+\frac{ \partial g}{\partial \bfx}(\bomu) \bfe
     $ 
     and applying linear Gaussian vectors transform yields approximately   $g(\bfx) \sim\mathcal (g(\bomu) ,\bfA\boSigma \bfA^T)$, where we let $\bfA:=\frac{ \partial g}{\partial \bfx}(\bomu)$.
     
     In the 2-sphere example of the present section,  our uncertainty representation may be taken as $\bfR_n=\exp(\boxi^\wedge)\hat \bfR_n$ with $\boxi\sim\mathcal N(\bfzero,\bfP)$, see \eqref{rr:eq} and Example \ref{ex:omega}. As a result it is rather easy to compute the covariance matrix of   $\bfR_n\bfL$ as follows. We may use linearizations to write that $\exp(\boxi^\wedge)\approx \bfI+\boxi^\wedge$ and thus 
      $
     \bfR_n \bfL=\exp(\boxi^\wedge)\hat \bfR_n\bfL\approx\hat \bfR_n \bfL+\boxi^\wedge \hat \bfR_n  \bfL=\hat \bfR_n \bfL- (\hat \bfR_n \bfL)^\wedge   \boxi=\hat \bfR_n \bfL+\bfA \boxi
     $ 
     with $\bfA=-(\hat \bfR_n \bfL)^\wedge$. As a result, the probability distribution of $\bfR_n\bfL$ is under a linear approximation  $\mathcal N(\hat \bfR_n \bfL,\bfA\bfP \bfA^T)$. 
     
     \section{Concluding Remarks}

If we step back a little and look  at the bigger picture, we see the main problem when designing filters on a manifold $\mathcal M$ is that we often lack coordinates to write down the filter equations on   $\mathcal M$. Even if we do, e.g. longitude and latitude on the sphere, 
this implicitly defines probability distributions on the manifold  in a way that may not suit the   problem well, see Fig. \ref{sphere:fig}. Over the past decades, researchers have advocated the   intrinsic approach based on the tangent space \cite{pennec2006intrinsic}. This way the filter becomes independent of a particular choice of coordinates on the manifold, but it   depends on the way tangent spaces at different locations correspond. Notably, we see at lines 5, 6, 7, 9 of Algorithm \ref{alg:1} the covariance matrix $\bfP^{+} $ is computed using local information at $\bchi$, in total disregard of $\bchi^{+}$, although   $\bfP^{+} $ is supposed to encode dispersion at $\bchi^{+}$! This means it is up to the user to define the way ``Gaussians" are transported over $\mathcal M$ from  $\bchi$ to  $\bchi^{+}$, as early noticed in  \cite{loiannoVisual2016}, see also \cite{svachaInertial2019}. The route we have followed herein consists in focusing on parallelizable manifolds where a global coordinate system of tangent spaces exists, and   readily provides a transport operation over $\mathcal M$.

However, there are multiple choices for the parallel transport operation. In  \cite{loiannoVisual2016,svachaInertial2019} the authors advocate  using the Levi-Civita connection for parallel transport, which depends on the chosen metric, and argue its virtue is that it is torsion free. In  the context of state estimation on Lie groups, though, the transport  operations that lead to the best performances are not torsion free, see    \cite{barrauInvariant2018}.  In  cases where it is unclear to the user which transport operation (in our case parallelization+retraction) shall be best, we suggest using our  code for quick benchmarking, as done in Figure \ref{fig:nav}. Indeed, the group structures $SE_2(3)$ versus $SO(3)\times \mathbb R^6$ actually boil down to  particular choices of parallelization (hence  transport), and the filter based on $SE_2(3)$ outperforms the other.

\bibliographystyle{ieeetr}
\bibliography{IEEEabrv,biblio,biblio2}

\end{document}